\documentclass[conference]{IEEEtran}
\usepackage{graphicx}
\usepackage{verbatim }

\begin{document}

\title{Dynamic Knowledge Graphs as Semantic Memory Model for Industrial Robots}

\author{\IEEEauthorblockN{Mohak Sukhwani,
Vishakh Duggal, Said Zahrai}
\IEEEauthorblockA{ABB Discrete Automation and Robotics\\
mohak.sukhwani@in.abb.com,
vishakh.duggal@in.abb.com,
said.zahrai@de.abb.com}}

\maketitle

\begin{abstract}
In this paper, we present a model for semantic memory that allows machines to collect information and experiences to become more proficient with time. Post semantic analysis of the sensory and other related data, the processed information is stored in the knowledge graph which is then used to comprehend the work instructions expressed in natural language. This imparts industrial robots cognitive behavior to execute the required tasks in a deterministic manner. The paper outlines the architecture of the system along with an implementation of the proposal.
\end{abstract}

\IEEEpeerreviewmaketitle

\begin{IEEEkeywords}
cognitive, intelligence, robots, co-bots, ontology
\end{IEEEkeywords}

\section{Introduction}
With latest advances in artificial intelligence (AI) it has been shown that the usability of machines is enhanced when they become capable to process data, find patterns and suggest proper actions. From predictive maintenance to improved Human-Machine collaboration, different intelligent systems are currently in use and facilitate our daily tasks. AI based techniques for modeling, forecasting and predicting machine behavior play crucial role in present day Industrial setups. With advancements in industrial connectivity and availability of rich sensory data, AI based techniques can achieve noticeable improvements in both efficiency and flexibility~\cite{fryman2012safety, RSsafe, mandal2019improving}.   

Industrial robots are machines that are made to perform complex tasks in factories. A typical robotic application~\cite{gruver1984industrial, lozano1983robot, biggs2003survey} requires a specialist to analytically decompose the complicated task into smaller sub-tasks and actions. The expert expresses detailed instructions in the form of robot programs to make robot accomplish desired goals. This process needs high level of expertise and is time consuming. An alternative approach is to use predefined set of templates~\cite{long2016template} for various tasks, prepared by experts - Block Based programming interface~\cite{weintrop2018evaluating} simplifies the process to some extent without any accuracy loss and makes it easier for non-specialists to learn robot programming. Although these methods help, they do not remove the needs for detailed instructions to be given to the robot every time a new task is to be executed.

\begin{figure}
  \centering
  \includegraphics[width=\linewidth]{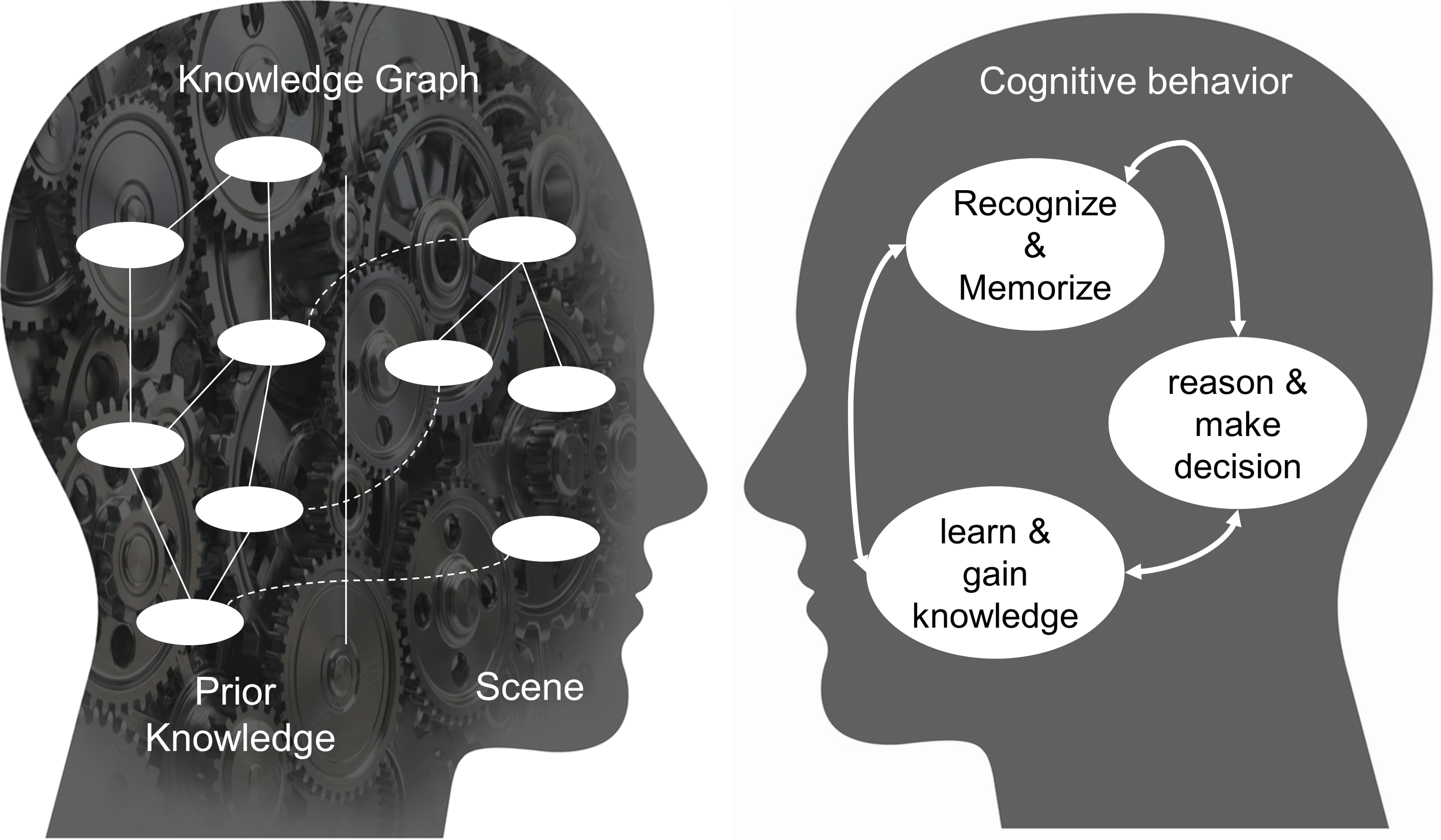}
  \caption{Cognition is the ability to acquire and process the information to learn various tasks necessary for living. Part of this information containing rules and facts is stored in Semantic Memory. The model proposed here aims to enhance learning ability of industrial robots. The proposed model is implemented using a dynamic knowledge graphs, where nodes describe the concepts and edges the interactions between the nodes.}
  \label{fig:CognitiveBehavior}
\end{figure}

Considerable effort in research societies and industry is put on making industrial robots more intelligent and thereby more useful. Most of the current work is based on use of neural networks and especially deep learning methodologies. These methods allow the robot learn motor skills in a way close to human learning. This make robots more skillful but for specific purposes~\cite{dnn_robotics} and ~\cite{rl_robotics}. There are other statistical methods that have been used to make robots more capable.~\cite{schaal2002scalable} uses locally weighted learning (LWL) method to learn complex real-time robot tasks. They introduce several ~\textsc{LWL} algorithms to model complex phenomena during autonomous adaptive control of robotic systems.

A useful and smart robot needs to cover large spectrum of applications. As the application areas steadily grow, the systems must accumulate experience, supervise themselves and run diagnosis to avoid mistakes and errors. In an industrial environment, they need to process a work instruction, extract information and execute requested tasks consistent with valid context in the environment, prior collected experiences and rules extracted from product manuals. We claim that reaching a high level of generality, needs more elaborate memory model that are closer to human memory, see figure~\ref{fig:CognitiveBehavior}.

\begin{figure}[h]
 \centering
  \includegraphics[width=\linewidth]{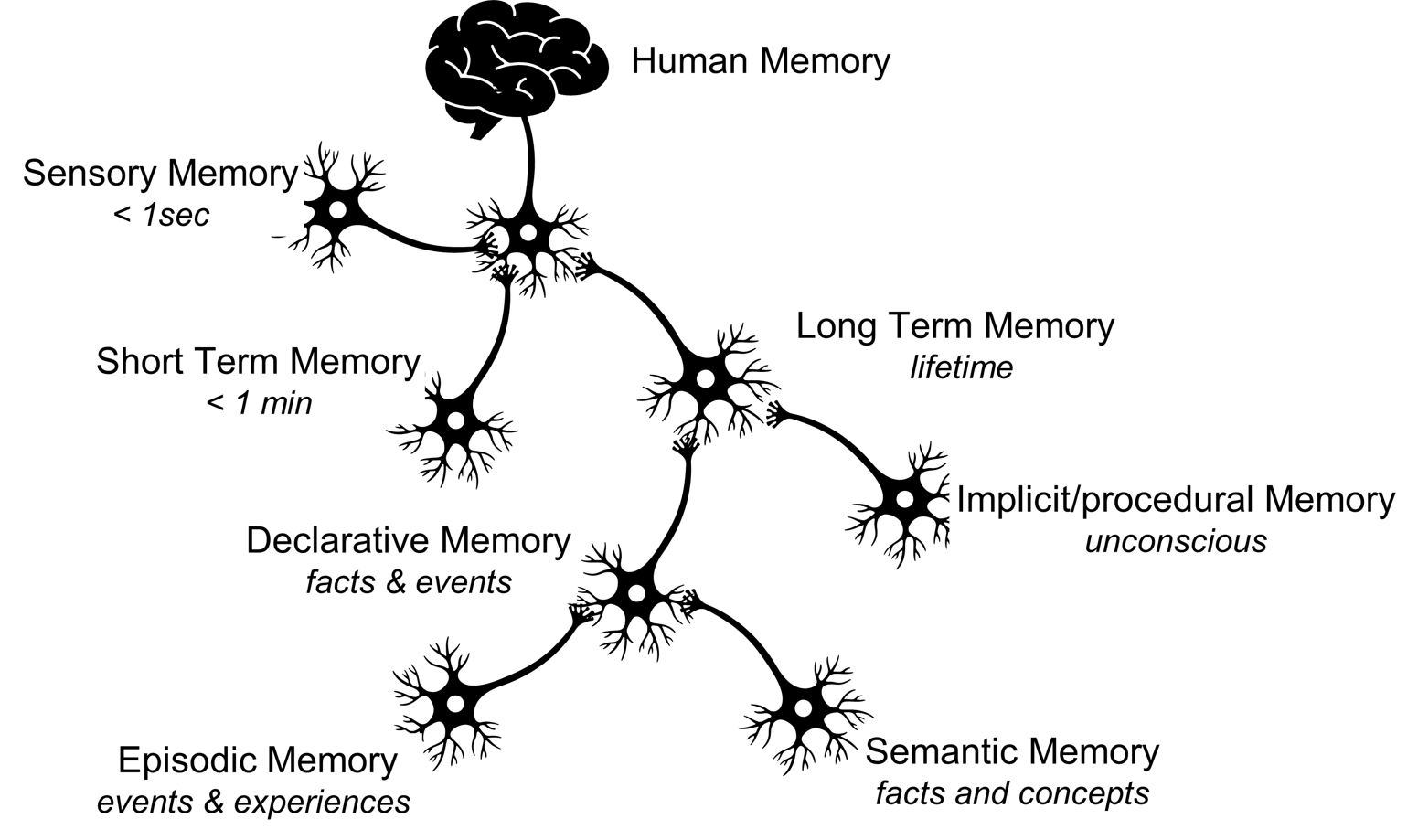}
  \caption{Generally accepted Human memory structure. Cognitive intelligence for industrial robots can be obtained if a form of declarative memory is conceived that allows the system collect experience, make decisions and use the skills.}
  \label{fig:HumanMemory}
\end{figure}

According to neuro-scientists, human long-term memory can be divided to procedural and declarative. Procedural or implicit memory is the unconscious memory that is used for learning skills, such as balancing, biking and playing musical instruments. These skills are typically acquired through repetition and practice. Declarative memory is where facts and events are stored and can be recalled consciously. Declarative memory is further divided into episodic and semantic memory, figure~\ref{fig:HumanMemory}. Episodic memory stores information about events, while semantic memory accumulates facts and data together with logical relations~\cite{h-mem}. The purpose of the work reported here is to build a model for semantic memory that allows the robot to relate current observations to prior ones, draw conclusion and take actions. Such a memory system is similar to an expert system, that could interact with all sensors and actuators and also dynamically expand. 

Regardless memorizing and learning capability of the robot, as any other work force in a factory, robots need instructions to operate in correct manner. In order to accomplish the goal, the robot needs to comprehend the instructions and perform complex sequences of sophisticated and accurate manipulation activities. As the instructions are given by a human, the most intuitive interface would be natural languages. However, in normal conversations concepts are vaguely described and statements and commands are often ambiguous and incomplete. If the instructions are expected to be given as natural language, we need to build a system that resolves the ambiguities in the language. 

The problem considered is to give a robot the ability to process a work instruction for an operator, extract information and execute the requested tasks. The robot is expected to execute tasks consistent with valid context in the environment, prior experiences, rules extracted from product manuals and the data collected by all interfaces and sensors available to the system. We aim to do so by semantic analysis of the valid context in relation to the prior knowledge and resolve the discovered ambiguities. New and unseen concepts, if any, are to be made clear in conversation with human operators. The newly defined concepts should be accumulated in a semantic memory representation that allows expansion of the capabilities of the robot. We believe that dynamic ontology-based frameworks are strong alternatives to implement such a memory system. We use Natural Language Processing (~\textsc{NLP}), computer vision and knowledge graphs to demonstrate the effectiveness of our approach on ABB's, IRB 14000, YuMi.

It is worth pointing out that our focus is on semantics of robotic manipulation in a manufacturing environment and we do not intend to cover the world knowledge. The system needs to recognize a set of objects and instructions that are related to manipulation of them, limiting the scope of the problem to object-action semantics for manufacturing purposes. The memory model proposed here is not an alternative to other related machine learning methods, but a complementary function to enhance the learning ability. 

\section{Related Work}

With affordable computing power, use of AI for industrial applications has become a realistic target and a subject of academic research and industrial development. Robot Programming languages and systems~\cite{gruver1984industrial, connolly2009technology} have been developed to instruct the robot move accurately in a predefined manner or adapt to the environment by reacting to sensory inputs. Commercially available robot systems~\cite{robotics2019yumi, robotics2014technical} provide that with variable burden on the user; the more accurate the robot system needs to be, the more detailed instructions are to be given and programming becomes more cumbersome. Interestingly, in many of such complex cases, a human operator does not even need any instruction. For example, programming a robot to perform cable assembly is a tough engineering work, while instructing an operator can be as easy as making a circuit diagram available. Experienced workers know how to read the diagram and perform the task and inexperienced ones can either learn very fast or simply find out by experimentation in a very short period of time. The question will then be, to what extent AI can reduce the need for detailed programming and allow robots be instructed as human workers ?

Expert systems were perhaps the first form of artificial intelligence used for industrial automation. They are memory based approaches that comprise of knowledge database and an inference engine. Expert systems have been used in past to solve problems like fault detection and fault tolerance, path and trajectory planning, obstacle detection, robot control, path planar optimization for industrial robotics~\cite{meijer1991robot, kishan2012review,kaldestad2012cad}. The function of semantic memory is very similar to an expert systems with extended ability to handle general sensory information and be dynamically expandable.

With availability of big data sets, machine-learning based robotic solutions have gained considerable traction. Supervised learning in vision AI for identification and sorting of objects~\cite{xia2016workpieces, ali2018vision} and vision-based pose estimation~\cite{yoon2003real} are example of use of AI for robot perception. Unsupervised Learning approaches have also been used in industrial settings to augment safe human robot collaborations~\cite{RSsafe, mandal2019improving}. End-to-end systems that learn robotic grasping~\cite{dexnet, dexnet2, dexnet3} can now assist robots to pick up the parts that it has never seen before. These methods were formulated as learning policies for bin picking of multiple objects in robot's field of view. Similar work has been done by using self-supervised learning~\cite{berscheid2020self, berscheid2020self} for precise pick-and-place application without explicit object models, which has drastically reduced the training time. Bayesian or probabilistic models are used for Imitation Learning~\cite{muhlig2009task, chella2006cognitive}, where the robot learns to perform a task from demonstrations. 

Although natural language processing is not a new area of science, with advances in machine learning techniques, the concepts have been elaborated fast during the past years. Semantic role labeling,~\textsc{SRL} is a technique of shallow semantic parsing of natural language text to generate predicate-argument structures of sentences. Shallow Semantic structures play a crucial role in natural language understanding as they abstract syntactic and morphological representations of sentences.Predicates generally are verbs and verbal nouns. They define the core meaning of the situation expressed in the text. Arguments on the other hands are phrases that answers questions as `who?', `where?', `when?' etc. They define the essential details to the situation expressed by a predicate. End-to-end deep models~\cite{zhou2015end} for~\textsc{SRL} seem to work well without syntactic inputs. Traditional Long Short Term Memory,~\textsc{LSTM}, based architectures make sequential prediction for arbitrary length sentences. They treat each sentence as sequence of words and recursively process them. Present state of art methods~\cite{he2017deep, palmer2010semantic} use deep attentional neural networks for the task of~\textsc{SRL}. When compared to traditional Recurrent Neural Networks~\textsc{(RNN)}, they draw the global dependencies of the inputs using deep highway bidirectional~\textsc{LSTM}s with constrained decoding.

Probabilistic Action Cores~\textsc{(PRAC)} interpreter~\cite{PRAC1,PRAC2} for natural-language instructions resolves the vagueness and ambiguity of natural language. The system infers missing information pieces to render an instruction executable by a robot. It formulates this as a reasoning problem from the perspective of Bayesian cognition. The system uses Markov logic networks and learns joint probability distribution over semantic network, to generate specific semantic representation and infer missing aspects of action specification. It is connected to a robot planner that executes the computed plans. Robotic agents have plan library that has parametrizable plans for actions. The actions are refined and instantiated according to the instructions generated. Probablistic Graphical models are constructed for each relevant action verbs and the most probable action cores are inferred. The action roles attached to respective action cores are retrieved from~\textsc{PRAC} knowledge base.

Ontology provides a framework for representing knowledge in a specific domain and assists in understanding a specific attribute of the world. Traditionally ontologies have been used to make web intelligent (semantic web). It captures the structure of the world and describes the domain knowledge. There are no strict and correct ways to describe domain and create ontology. As a standard framework people have used Resource Description Framework~\textsc{(RDF)} and Web Ontology Language~\textsc{(OWL)} to create ontologies for high expressiveness~\cite{olszewska2017ontology, paull2012towards}. 

\section{Semantic Memory Model}
In this section, we explain the basic ideas of our memory management system and roughly how the memory manager interacts with the devices present in the system to gather relevant information, store/expand the knowledge and execute correct actions according to instructions. Our focus is robotic manipulation in manufacturing environment following the given set of instructions. 

\begin{figure}[t]
 \centering
  \includegraphics[width=0.9\linewidth]{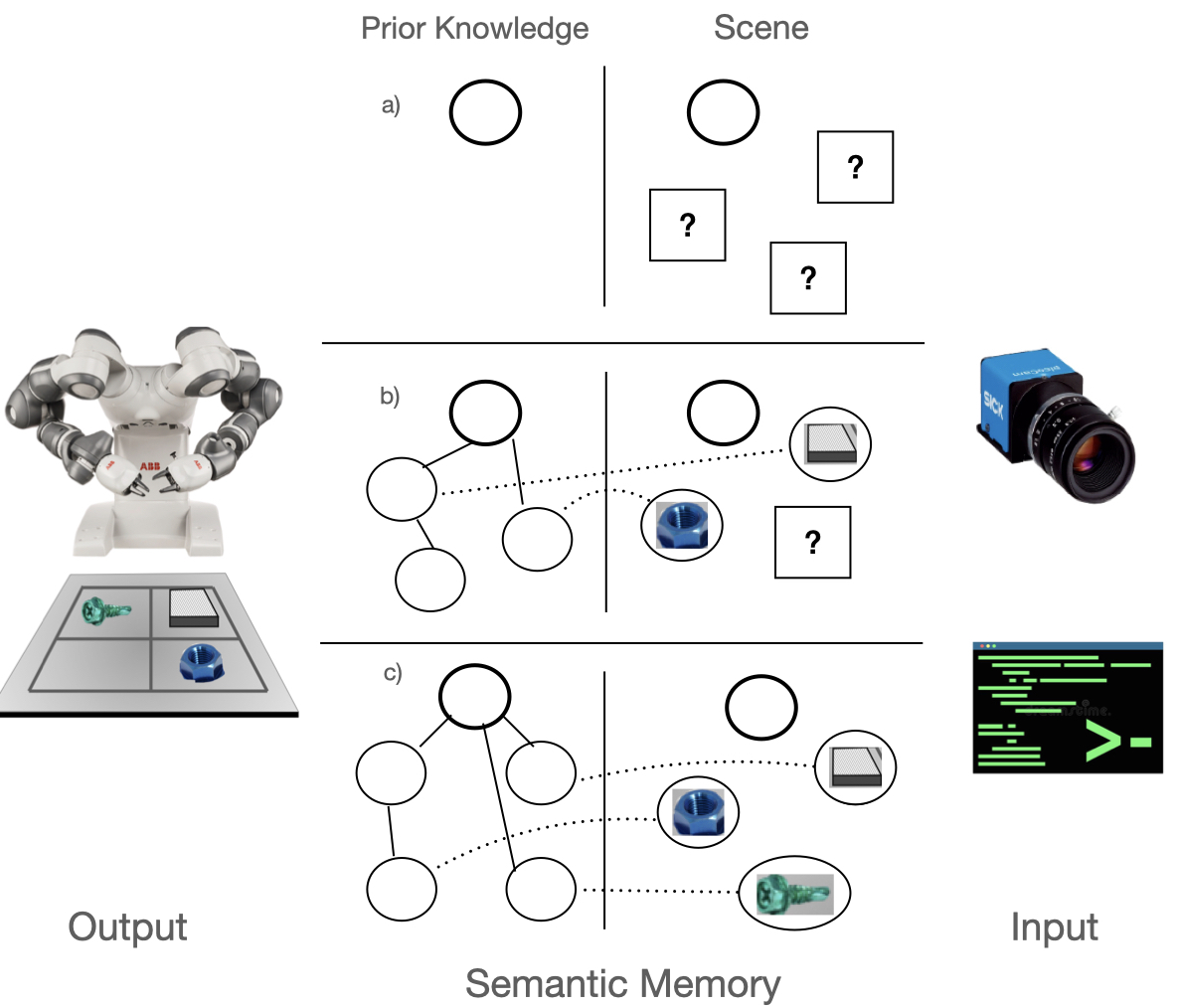}
  \caption{Output and input devices are presented on left and right side, respectively. The main actuator is the robot, and information is collected by sensors like camera or command interfaces. The semantic memory as a knowledge graph is in presented in the the middle, where  objects are marked with square boxes and "?" before they are recognized and thereafter with circles. (a), (b) and (c) show consecutive stages of knowledge graph explained in this paper.}
 \label{fig:system}
\end{figure}

For a system to be able to interact with the surrounding and manipulate reachable objects, it needs to sense and recognize the objects in the scene. Recognition in practice is a successful match between measured descriptive quantities and preexisting descriptions. Learning can be understood as accumulation of information in an organized manner. The ability to reuse of that in a controllable manner, to achieve desired effects, can be considered as knowledge. We propose a system that that is able to perform these steps and thereby provide necessary ability to industrial robots to become steadily more competent over-time to perform more manufacturing tasks and in a better way. 

\begin{figure*}
\centering
  \includegraphics[width=0.9\linewidth]{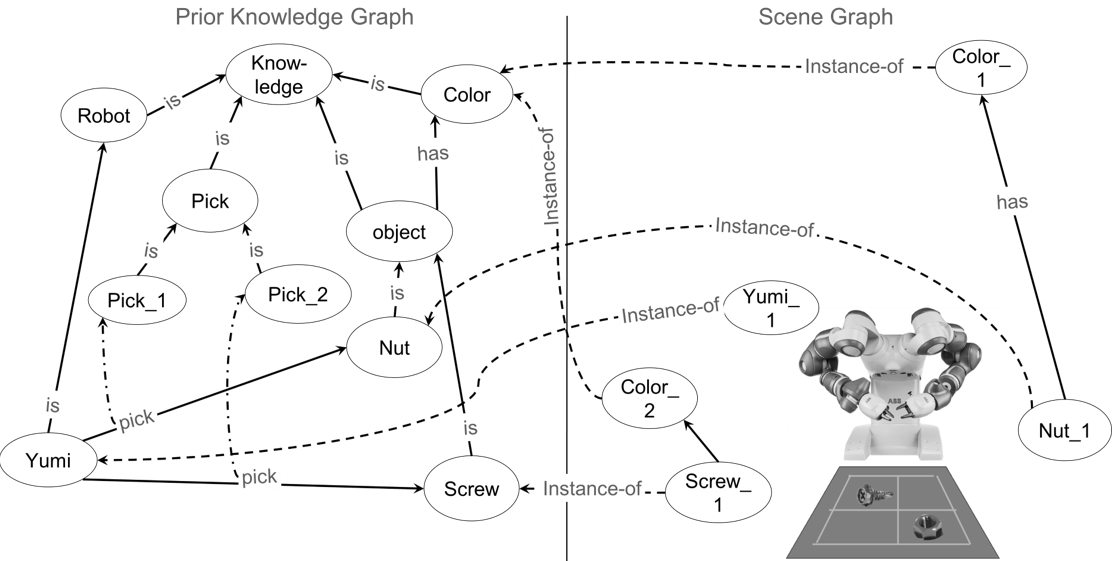}
  \caption{A semantic memory model to store information and use the experiences in a deterministic manner. We observe a subset of active elements of prior knowledge and scene graphs with associated relational links between the them.}
  \label{fig:SemanticMemory}
\end{figure*}

The system consists of three essential components: (1) input devices and sensors (2) output devices and actuators (3) the semantic memory module. Memory module organizes and stores the collected information from input devices. It commands the output devices such that a realistic interaction with the environment can be possible. We assume that there is an input device to receive commands in natural language and that there is a vision system to be able to sense the surrounding. The robot system, with its conventional controller, is in fact a collection of actuators and sensors. However, at this stage, we consider robot as an actuator that is expected to manipulate objects. Further, we focus on memory management and assume that each device is smart enough for the interaction with the semantic memory manager.

The core of our semantic memory manager is a dynamic knowledge graph, where the nodes represent concepts and objects and edges define relations between two nodes or how they interact with each other. The meaning of the edges, or the related action, is defined by another node in the graph. The graph initially has only a minimum number of concepts, such as elementary geometries like sphere, cube and basic relations like `is', `has' and `is instance of'. The graph grows in interaction with devices connected to the system and what they observe or do. Conversation with a human operator will be necessary for defining new concepts in the system.

The knowledge graph has two distinct parts that are tightly interconnected. These are the ~\textit{scene graph} and ~\textit{prior knowledge graph}. The scene graph describes the surrounding world, while the prior knowledge graph stores all concepts that have been defined earlier. One can argue that the prior knowledge is type definitions while the scene graph holds information of objects that are instances of those types.

Once an object is identified in the scene by one of input devices, its properties are measured and stored. Assuming that the input device is a camera, once an object is detected, the camera is expected to output at least approximations for location of the object, its shape, size and color, so that the object can be identified. This means that the camera is smart enough for our purpose. The measured properties are compared with the existing types and concepts. If a match is found, the identified object is labeled by creating relations to relevant nodes of the prior knowledge graph with information about the type of the object. If no match is found, a new concept will be defined and will be labeled by requiring information from the operator. Obviously, with AI-based image processors, a large number of objects can automatically be labeled, without help from an operator.

\begin{figure}[t]
\centering
  \includegraphics[scale=0.45]{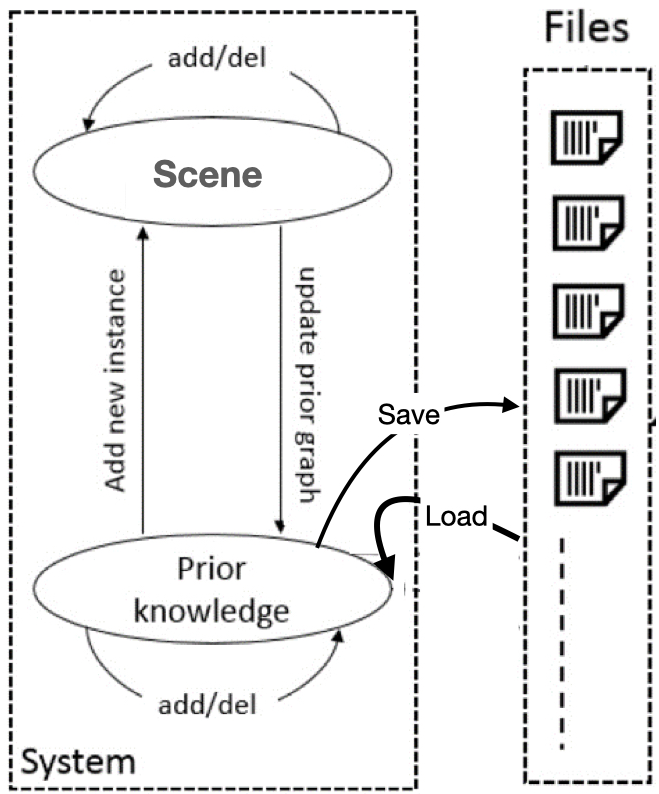}
  \caption{Schematic view of the memory management architecture, interaction between accumulated knowledge and scene view and persistent storage for the prior knowledge.}
  \label{fig:SystemArchitecture}
\end{figure}

As mentioned above, the robot, say ABB's IRB 14000, also called Yumi, is an actuator that operates on operands according to the instructions. Instructions are treated in exactly same manner as an observed object, as explained above. The sentence is recorded by the input device and is analyzed. Once the roles in the sentence are defined, the prior knowledge graph will be searched to find the relevant action or start to record a new one. Execution of instructions means that the robot operates on an object. For the topic of interest, it is not important how the operations are implemented or learnt. They can be programmed in traditional manner, be implemented with some kind of learning method or can be taught the first time the instruction is given and are in general dependent on the type of the operand. A robot system with such an ability is smart enough for our purpose.

Figure~\ref{fig:system} presents an example of the knowledge graph, the system and how the instruction are received. At starting point, the prior knowledge graph contains a number concepts. The vision sensor records the scene and for each object defines its signature so that the corresponding type can be found and associated to the object. These associations are shown as links between elements in the scene graph and elements in the prior knowledge graph. 

Figure~\ref{fig:SemanticMemory} shows a subset of the prior knowledge graph related to the actions. Here, we see that three different `Pick' actions are defined for three different objects. At this moment, if the instruction will be `Yumi, pick the nut!' the system will choose pick1, and look for a nut in the scene graph, get its locations and execute the action with those inputs. If a new object will need to be picked, Yumi will be able to suggest to take it as a nut, screw or a clip and therefore no new programming or teaching will be needed if the existing actions are sufficient. The instructions can be more specific with for example specifying the color of the nut. In that case, a nut with the specific color will be sought in the scene graph. 
The strength of semantic modeling can be observed here. The system does not learn an instruction as a whole, but learns each concepts separately and combine them to take the right action. How exactly it is done will become more evident by description of the implementation details and analysis of the experiments in the two following sections.

\section{Implementation}
In this section, we present the details of our current implementation. We are building a semantic memory and intend to use it as a knowledge base for robotic manipulation. The complete cycle includes data collection from environment and its analysis, storage in the graphs and its use for execution of instructions, figure~\ref{fig:SystemArchitecture}

\subsection{Semantic Data Processing}\label{section:vision}
\begin{figure}[b]
  \centering
  \includegraphics[width=\linewidth]{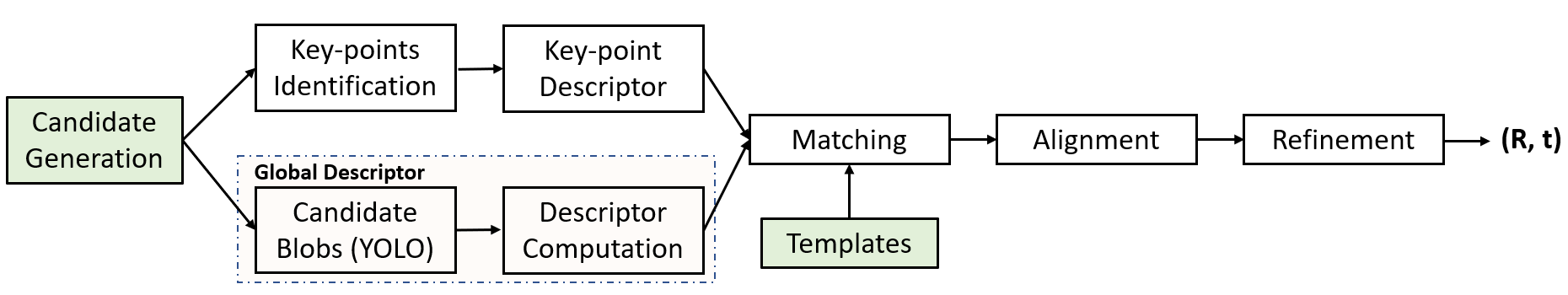}
  \caption{Workflow for estimation of object coordinates (x,y,z) and object orientation (yaw,pitch, roll).}
  \label{fig:6DoF}
\end{figure}

For a robot to execute a requested instruction and operate on objects, it needs to analyze the inputs from all sensors together with the instruction itself. We consider all input devices together and intend to provide a semantic analysis that allows finding correct objects and actions.

Identification of object, finding its localization, orientation, color, distance from the image sensor etc. are some pieces of information that are needed for proper operation. We use state-of-the-art, real-time object detection~\cite{huang2018yolo} based approach followed by 3-D template matching to detect object coordinates (x,y,z) and compute object orientation (yaw,pitch, roll), see figure~\ref{fig:6DoF}. The system  detects objects in semi-moderate complex environment, viz. homogeneous background with modest illumination variations. The  approach is invariant to shape, size, texture and other properties of objects.

We use NVIDIA® Jetson Rudi Embedded System board~\cite{mittal2019survey} with 1 TFLOP/s, 256 CUDA cores, 8GB LPDDR4 memory, 64-bit ARM® A57 CPUs (6 Core), 32GB eMMC for object pose detection.At heart of $3D$ object recognition and pose estimation is identifying set of correspondences between two different clouds, one of them containing the actual object and other containing the object from actual scene. We use global descriptors to understand the notion of object. The segmented point cloud from detected objects help us identify prospective point cloud clusters. We remove false positives by discarding point clouds smaller than certain threshold - the threshold is computed by observing size of objects in the database. We use, Clustered Viewpoint Feature Histogram,~\textsc{CVFH}~\cite{aldoma2011cad} based global descriptor to compute point cloud descriptors. Once all the descriptors have been computed nearest neighbors search is performed in the database to do correspondance matching, figure~\ref{fig:6DoF}. Viewpoint information encoded in descriptor is used to compute rotation and the roll angle is computed using, Camera Roll Histogram,~\textsc{CRH}. The final angle is computed by aligning the current~\textsc{CRH} with the stored descriptor.~\textsc{CVFH} is the fastest of all global descriptor among all the investigated descriptors. It takes 5ms to compute~\textsc{CVFH} descriptor without down-sampling the point cloud. With down sampling, the descriptor extracting time is decreased to 1ms.~\cite{pcl, pcltutorial}.
\begin{figure*}[h]
\centering
  \includegraphics[width=0.95\linewidth]{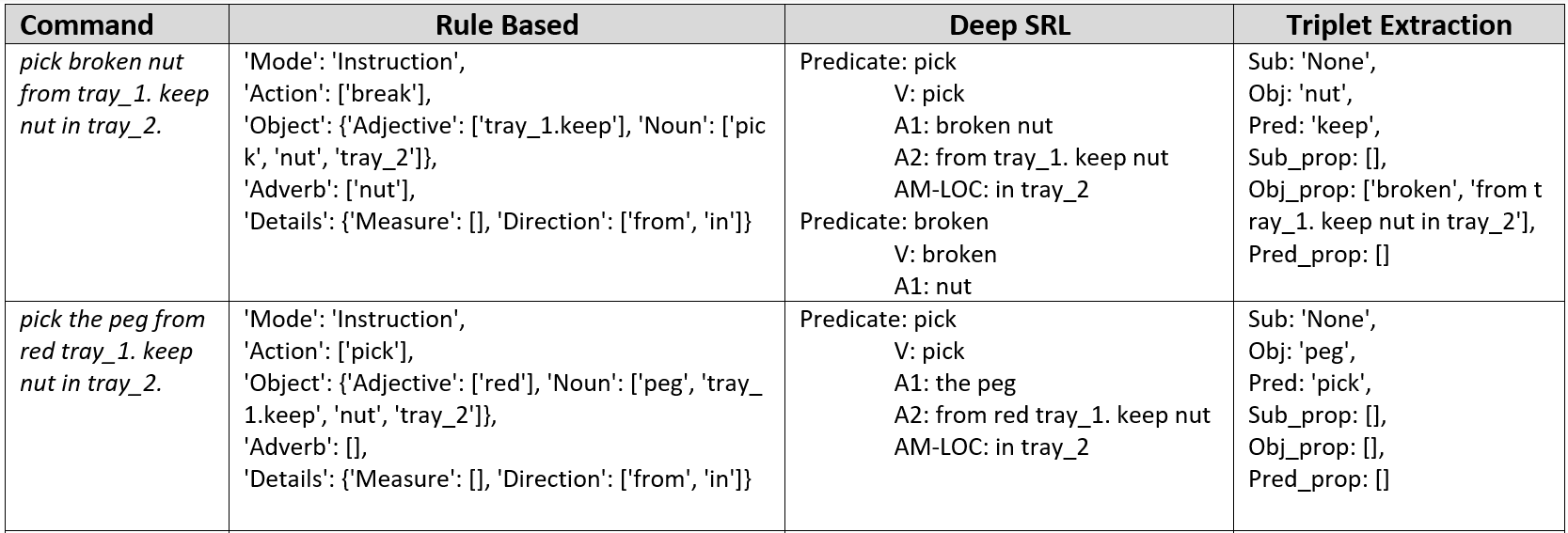}
  \caption{Comparison between different methods for understanding instructions using Heuristic over Part-Of-Speech (POS) Taggers, Deep SRL based method and subject-predicate-object triplet extraction.}
  \label{fig:nlp}
\end{figure*}

To process the instructions, we use deep~\textsc{SRL}~\cite{he2017deep} based approach to extract intent from user commands.~\textsc{SRL} determines the latent predicate argument structure of a sentence and provides representations that identify the meaning of sentence and relationships among the actors and verbs in the sentence. To baseline our approach we use Part-Of-Speech~\textsc{(POS)} Taggers to generates intermediate tags and use heuristics to generate formatted output. Figure~\ref{fig:nlp} compares both these approaches along with subject-predicate-object triplet extraction method~\cite{triplet}. Tuning such systems could be difficult and would require considerable efforts to cover all possible use cases. The output of intent extraction step is used by `dynamic' robot world knowledge ontology. 

\subsection{Building knowledge graphs}
The knowledge graph was implemented using graph-tool \cite{pygraph}, a Python 3 module for manipulation and statistical analysis of graphs or networks, which supports visualization of graph along with efficient on-the-fly filtering of vertices and edges including standard topological algorithms. Algorithms specific to our application along with interfaces for manipulation of the graphs were added through the API's provided by the library.

As mentioned above, the knowledge graph is divided into two sub-graphs, prior knowledge graph and scene graphs, each of which are populated by nodes interconnected with edges. Each element is an instance of a class and embeds both information and functions. Necessary functionality for information management, retrieval as well as exchange between the two sub-graphs are implemented in the head node of each, while other nodes and edges include graph manipulation operations such as creation, deletion, modification, providing specific instructions to be executed depending on the type of nodes and edges.  

Scene graph and prior knowledge are self-contained graphs, but dependent on each other to shape a complete understanding of the information. Initially both are empty, but the prior knowledge can be populated programmatically or by reading from file. The scene graph, which represents current active world, is populated by the sensory data in relation to the prior information available within prior knowledge graph. The steps are presented below.

\begin{figure}[b]
\centering
  \includegraphics[width=0.99\linewidth]{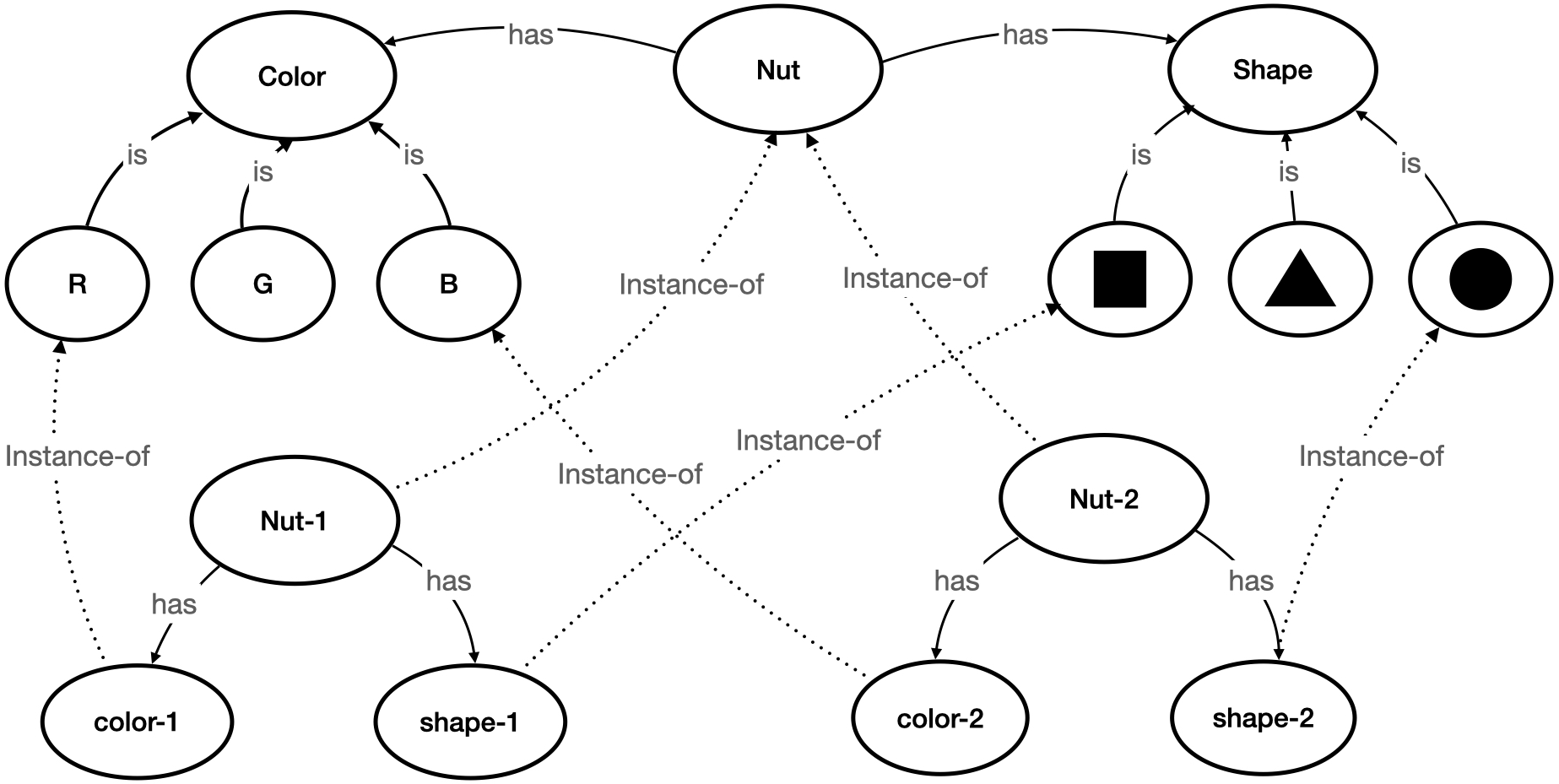}
  \caption{A general description of nodes and edges in semantic memory design. Nodes with static information represent either an object or properties. Edges define relation between source and destination nodes. 'R','G' and 'B' in graph denote red, green and blue colors respectively.}
  \label{fig:NodesAndEdges}
\end{figure}

Nodes in the graph hold two types of information: labels and properties, which either are static or dynamic. Node with static information represent either an object or properties. For example, the color node would store `Red' or `Green' and position node would store the position of object in real world (X=10,Y=55,Z=-10). Dynamic information corresponds to executable code, model or objects representing learned task, figure~\ref{fig:NodesAndEdges}.

The edges are directional and define relation between source and destination nodes. They can either be of predefined type, `has', `is' and `instance-of', or any user defined type, figure~\ref{fig:NodesAndEdges}. The predetermined types of edges have special importance in the graph as follows: 
\begin{enumerate}
	\item has: Source node has destination node as a property.
	\item is: Source node is of type destination node or child node of type destination node. 
	\item instance-of: Source node in scene graph is an instance of destination node in prior knowledge graph. This is the only relation or edge which spans across both the graphs.
\end{enumerate}

The user defined edges depict task with the given label. For example, `pick', `place', `test' are relations between the robot and an object and point to nodes containing the actual task implementation. The process to add an object to the scene graph with reference available in the prior knowledge graph: 

\begin{enumerate}
	\item Object type in scene is detected from sensory data along with its properties such as color, shape, size.
	\item Node matching the object is searched in the prior knowledge graph.
	\item A node is added to the scene graph and is linked with "instance-of" edge with selected Node.
	\item The logic embedded within `instance-of' edge renames the node to `Label\_instance\_number', where `Label' is the type of the object found and the `number' is the instance number. Scene graph does bookkeeping of number of instances of each Node.
	\item A list of all nodes linked to the found node in the prior knowledge graph with "has" edge is created.
	\item Each node in the list is added to the scene graph and attached with the similar structure as in prior knowledge, creating an exact copy in scene graph. This and all previous steps are repeated for all nodes being added to scene graph. Thus, all nodes in the scene graph will be linked to their type nodes in prior knowledge graph.
	\item Properties values collected from sensory data are added to the respective nodes in the scene graph.
\end{enumerate}
If no prior reference is found, additional information will be required to create relations to existing information or expand the prior knowledge.

\subsection{Execution of instructions}
When an instruction is given, it is broken down into parts with roles identified and is expected to be executed in context of current environment with the available accumulated knowledge. To do that, nodes within scene graph and the prior knowledge graph are matched with respective parts in the instruction. This is done in the following way:
\begin{enumerate}
	\item For each named object, e.g. `Yumi' and `nut', the respective node in the prior knowledge graph is found.
	\item For each named object, a list of all nodes connected to the respective node in prior knowledge graph with ``is'' and ``instance-of'' is created.
	\item If there are any properties of the object requested in the instruction such as color or size, the list will be filtered accordingly. 
	\item When a set of objects are now identified, the right action with the given label is searched in prior knowledge graph as an edge to link the right actor to the right object.
	\item With all elements in the instructions identified, the instruction can be executed.
\end{enumerate}

The above steps will become clearer with an example. Assume the instruction ``YuMi, pick the nut!''. This is first broken into `YuMi', `pick' and `the nut' and respective nodes in the knowledge graph are identified. Thereafter, the objects in the scene graph that relate to right types are identified; this will result in `YuMi-1' for `YuMi' and `Nut-1' and `Nut-2' for `Nut'. The list will be filtered if for example a specific color is requested. If there is a match the target is found, otherwise the closest match is proposed. For example, if Nut-1 had property color as blue, while green is requested, the system will inform the user about inability to find exact object and will ask whether the blue one can be accepted. Equally, for the action, an edge with label `pick' connecting `YuMi' to a “nut” is searched. If it is found, the action can be executed and if there is none, the closest one will be suggested. 

\begin{figure}[t]
 \centering
  \includegraphics[width=0.8\linewidth]{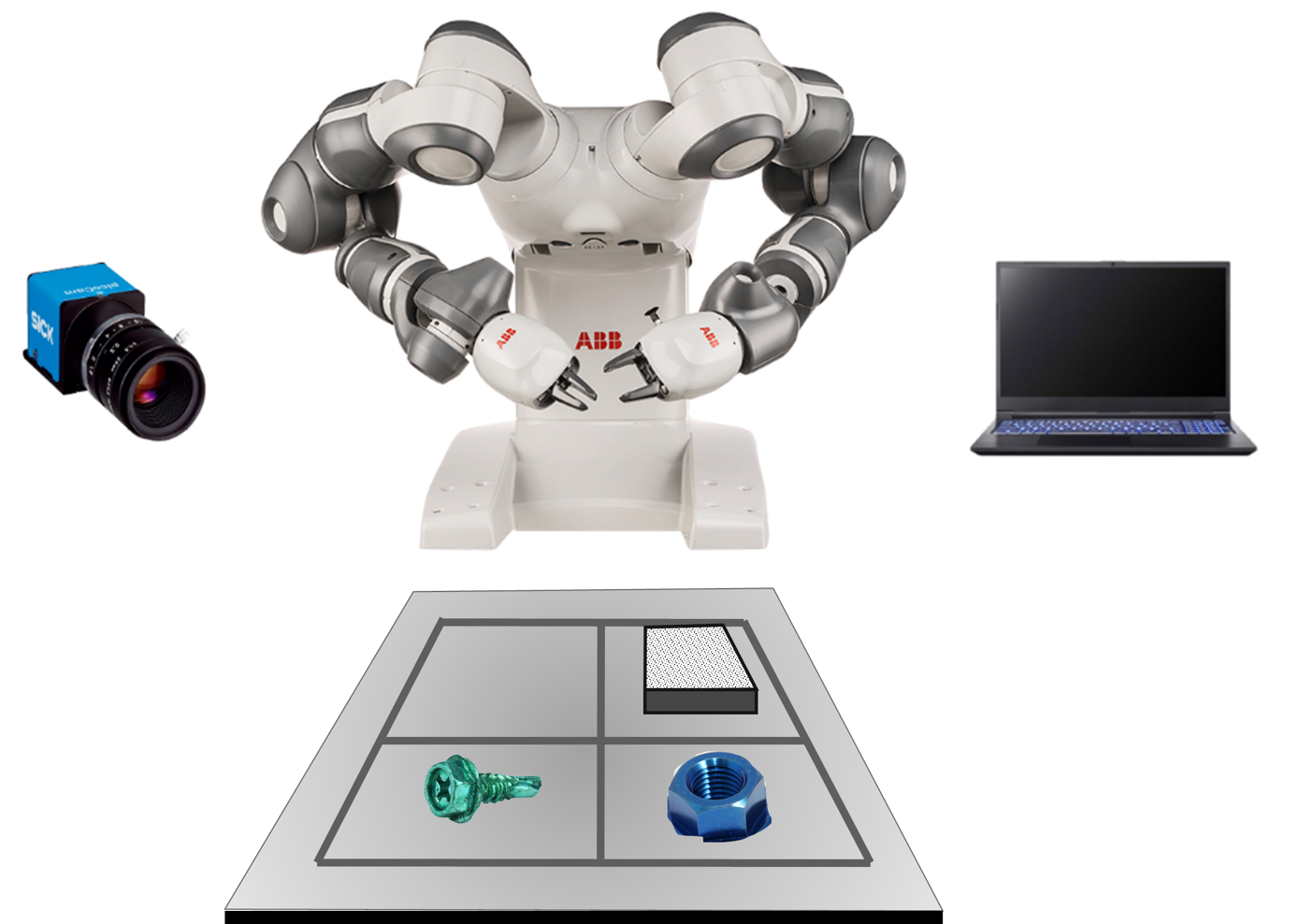}
  \caption{The experimental setup includes: 1) YuMi IRB 14000, 2) Computing device, a laptop, 3) a Vision sensor and 4) Industrial objects located on the table.}
  \label{fig:ExperiementalSetup}
\end{figure}

The above steps show how information is divided into knowledge that needs to be saved and scene information that should be built dynamically, how it is stored and how the information is used to execute an instruction. The system is complete if the necessary information is available. Further, it was explained how object information is stored in the scene graph and how a new type can be added to the knowledge graph. What remains is the definition of action. In general, we consider the action as a skill that the system should be capable of. Exactly how the skill is learned or programmed is not information that needs to be stored in the semantic memory and therefore not in the scope of this work. In the current system, we have used available interfaces in ABB’s controller for a convenient programming. The robot is moved from one position to another, simply by guiding it by hand and the gripper is activated by a command. The actions can of course be provided as a library of pre-programmed skills or as an AI-based skill package that should be trained.

\section{Experiments}

\begin{figure}[b]
 \centering
  \includegraphics[width=\linewidth]{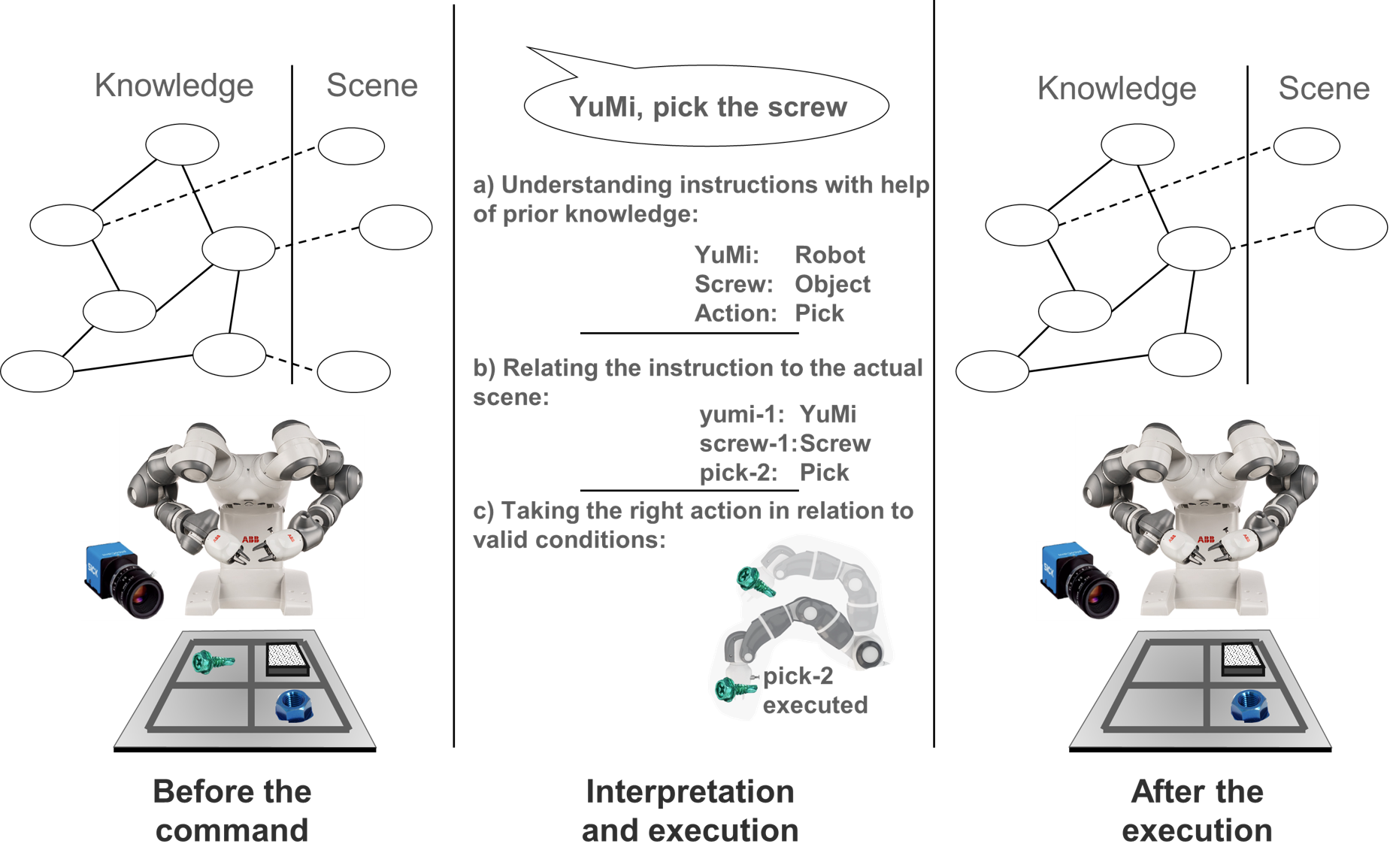}
  \caption{Interaction with system and execution of instruction. When the system starts, prior knowledge is loaded and the scene graph is populated with observed objects (left). The system acts as shown in the middle column: When an instruction arrives, it is analyzed and interpreted with help of prior knowledge (left). The interpretation is matched with the scene graph to find correct objects (middle). Once the objects are identified and proper action is found, the instruction will be executed. (right) The scene graph is updated, reflecting the changes in the environment.}
  \label{fig:Experiment1}
\end{figure}

A typical automation system includes, in addition to the robot itself, a number of other elements. Elements such as grippers, fixtures, sensors, trays etc. work in tandem to make a robotic application come to life. The objective of these experiments is to enable the robot to operate on the repeated tasks with least amount of programming effort. We teach a robot to learn by demonstration and reuse its experiences using the memory structure we propose in this work. Our experimental system, figure~\ref{fig:ExperiementalSetup}, is composed of the following elements:

\begin{enumerate}
	\item YuMi IRB 14000~\cite{robotics2019yumi}, dual arm collaborative robot with grippers .
	\item A table to perform experiments on objects with robot.
	\item Objects such as Nut, Box, Screw etc. kept on the table.
\end{enumerate}

We use vision algorithms, section~\ref{section:vision}, to detect object coordinates (x,y,z) and compute object orientation (yaw,pitch, roll) in semi-moderate complex environment. Intel Core i3 4590 based laptop with $4$GB RAM running on Ubuntu 18.04, was used to interface with Jetson Rudi Embedded System board~\cite{mittal2019survey}. We used Python3 to implement majority of the components of the platform. The robots used were programmed using ABB RAPID~\cite{robotics2014technical} language. 

We demonstrate the details of three experiments in the following sections. The details include robot interactions, command execution sequence, status of memory modules and update details of the respective knowledge graphs .    

\subsection{Experiment 1:"YuMi, pick the screw!"}
\begin{itemize}
    \item Sensory module detects three items on the table. The item details are added to the scene graph: `nut\_1' with color property `green' of type Nut, `screw\_1' with color property `blue' of type Screw and `box\_1' with color property `gray' of type Box and "yumi\_1" of type robot YuMi.   
    
    \item `yumi\_1' and `screw\_1' are selected and prior knowledge is checked. Prior knowledge has information about `pick' task with respect to YuMi and Screw. Hence, the task is executed and screw is picked and removed, as observed in figure \ref{fig:Experiment1}. 
    
    \item Sequentially, another command "YuMi, pick the green nut!" is given. Both `yumi\_1' `nut\_1' are are found in prior knowledge graph. The task information for ``pick'' with respect to YuMi and Nut and is executed. `nut\_1' is picked and removed from scene. 
    
\end{itemize}
In the above list, \_number denotes the sequence number of objects of the same type. For example, nut\_1 and nut\_2 are the first two instances of object type Nut.
	
\subsection{ Experiment 2: "YuMi, pick big clip!"}
\begin{itemize}
    \item Sensory module detects two items and adds them to the scene graph : `clip\_1' with color property `green' and shape `big' and `clip\_2' with color property `blue' and shape `small' of type Clip along with `yumi\_1' of type robot YuMi. 
    
    \item As there are two nodes are of type clip, the properties of each node are matched with the shape property `big'. Scene graph has information of `clip\_1' and `yumi\_1' and prior knowledge details of `pick' task with respect to YuMi and Clip. Hence, the task is executed and "clip\_1" is picked and removed. 
    
    \item Sequentially, another command "YuMi, pick the blue clip!" is given, `yumi\_1', `clip\_2' are selected as property `blue' is matched with `clip\_2' and task is executed.
	
\end{itemize}
	
\subsection{Experiment 3: "YuMi, pick the new\_obj!"}
\begin{figure}
 \centering
  \includegraphics[width=\linewidth]{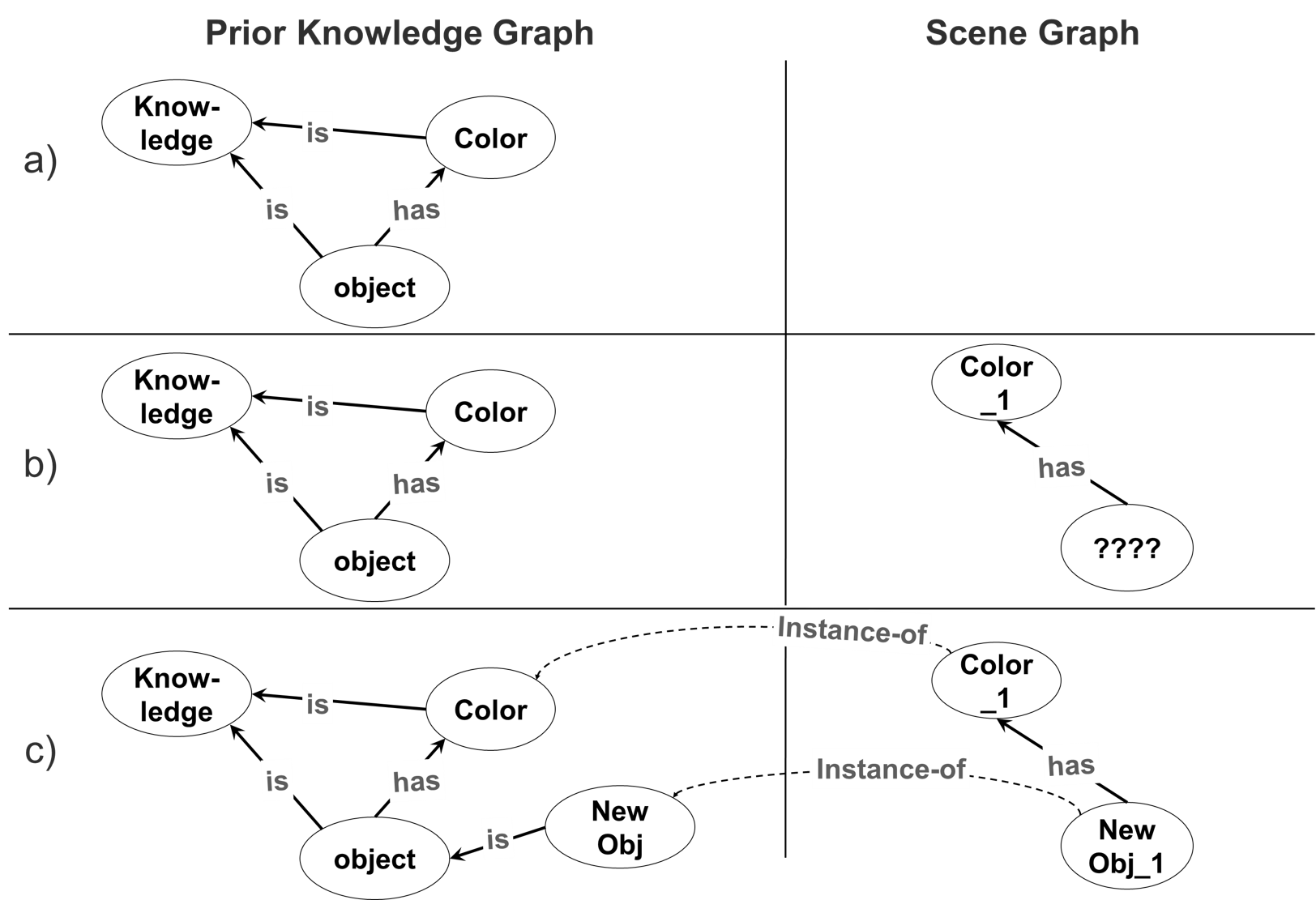}
  \caption{Learning from observations. Initially, the graph includes basic types and relations only (a). When a new object is observed, its features are quantified (b). After that the observation is labeled with human help, both the prior knowledge and scene graph are updated to include the new information (c).}
  \label{fig:Experiment3}
\end{figure}

Here, "new\_obj" is used to refer to a label that is not known to the system.
\begin{itemize}
    \item Sensory module detects two items and adds them to scene graph : `box-1' with color property `gray' and shape `square' and `new object' with color property `gray' and shape `square' along with `yumi\_1' of type robot YuMi. 
    
    \item The system, informs user about ``new object'' along with detected properties. User instructs the system to add this to prior knowledge graph. User either provides details about its relation to existing objects,i.e "box" or "Nut" etc or instructs the system to create a new type in prior knowledge. In case a new type is created, node `new\_obj\_1' is created in scene graph. 
    
    \item Since details of `pick' are also missing from the graph, system asks user to either train a new task (by demonstration) or link the same to one of the existing task labels based on matching properties of other nodes, figure \ref{fig:Experiment3}.
\end{itemize}
\section{Conclusion and Future Extensions}

Similar to humans, industrial robots can use a combination of memory models to analyze the environment, register and accumulate the processed information for later use. While general Machine learning based models are good for skill learning, we need semantic memory to learn long term innate concepts. Such memory models are similar to robotic expert systems which are extended to general sensory information. The approach presented in the paper, describes a memory management architecture to learn from complex environments and structure the same for future use. This structure when used in tandem with~\textsc{NLP} engine and vision cues leads to reduction in programming time and increase in productivity. 

We intend to extend the capabilities of intent extraction algorithms, so that they can deal with complex set of advanced human commands. This would require improvements in robot execution planners too. Input commands in form of human speech are also being experimented with. To make memory representations more robust and extensible, we foresee a future where base ontology can be initiated using web data and product manuals. The ability to forget experiences over a period of time and to be sure of the ones which robot experiences quite often are few of the memory management features we feel would be important in long run.  

\bibliographystyle{IEEEtran}
\bibliography{IEEEexample}

\vspace{1cm}
\hspace{-0.5cm}

\end{document}